\documentclass[10pt,twocolumn,letterpaper]{article}

\usepackage[table,dvipsnames]{xcolor}

\usepackage{cvpr}

\usepackage[accsupp]{axessibility}  

\usepackage{graphicx}
\usepackage{amsmath,amssymb,amsthm}
\usepackage{mathtools}
\usepackage{booktabs}
\usepackage{comment}
\usepackage{color}
\usepackage[nice]{nicefrac}
\usepackage{multirow}

\definecolor{fbApp}{HTML}{c8e7fa}
\newcommand{\cc}{\cellcolor{fbApp}}
\newcommand{\fcg}{\color{black}}

\usepackage[pagebackref,breaklinks,colorlinks]{hyperref}

\usepackage{amsmath,amsfonts,bm}

\def\eqref#1{equation~\ref{#1}}

\def\1{\bm{1}}

\def\vm{{\bm{m}}}

\def\vs{{\bm{s}}}

\def\vx{{\bm{x}}}
\def\vy{{\bm{y}}}
\def\vz{{\bm{z}}}

\DeclareMathAlphabet{\mathsfit}{\encodingdefault}{\sfdefault}{m}{sl}
\SetMathAlphabet{\mathsfit}{bold}{\encodingdefault}{\sfdefault}{bx}{n}

\usepackage[capitalize]{cleveref}
\crefname{section}{Sec.}{Secs.}
\Crefname{section}{Section}{Sections}
\Crefname{table}{Table}{Tables}
\crefname{table}{Tab.}{Tabs.}

\def\cvprPaperID{11971}
\def\confName{CVPR}
\def\confYear{2023}

\newcommand{\putalg}{{I-JEPA}\xspace}

\title{
  Self-Supervised Learning from Images with a\\ Joint-Embedding Predictive Architecture
}
\author{\normalsize\bf Mahmoud Assran$^{1,2,3}$\thanks{\texttt{massran@meta.com}} \quad\bf Quentin Duval$^{1}$ \quad Ishan Misra$^{1}$ \quad Piotr Bojanowski$^{1}$\\
\normalsize\bf Pascal Vincent$^{1}$ \quad  Michael Rabbat$^{1,3}$ \quad Yann LeCun$^{1,4}$ \quad Nicolas Ballas$^{1}$\\[1em]
\normalsize$^{1}$Meta AI (FAIR) \quad $^{2}$McGill University \quad $^{3}$ Mila, Quebec AI Institute \quad $^{4}$New York University}

\begin{document}
\maketitle

\begin{abstract}
This paper demonstrates an approach for learning highly semantic image representations without relying on hand-crafted data-augmentations.
We introduce the Image-based Joint-Embedding Predictive Architecture (\putalg), a non-generative approach for self-supervised learning from images.
The idea behind \putalg is simple: from a single context block, predict the representations of various target blocks in the same image.
A core design choice to guide \putalg towards producing semantic representations is the masking strategy; specifically, it is crucial to (a) sample target blocks with sufficiently large scale (semantic), and to (b) use a sufficiently informative (spatially distributed) context block.
Empirically, when combined with Vision Transformers, we find \putalg to be highly scalable.
For instance, we train a ViT-Huge/14 on ImageNet using 16 A100 GPUs in under 72 hours to achieve strong downstream performance across a wide range of tasks, from linear classification to object counting and depth prediction.
\end{abstract}
\begin{figure}[t]
    \centering
    \includegraphics[width=\linewidth]{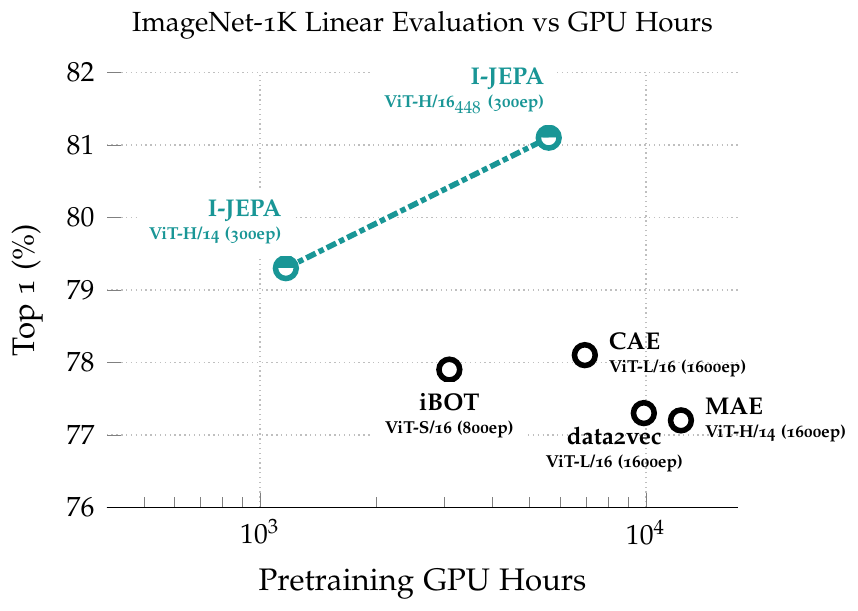}
    \caption{{\bf ImageNet Linear Evaluation}.
    The \putalg method learns semantic image representations without using any view data augmentations during pretraining.
    By predicting in representation space, \putalg
    produces semantic representations while using less compute than previous methods.}
    \label{fig:mae-comparison}\vspace{-1em}
\end{figure}

\begin{figure*}[t]
    \centering
    \begin{subfigure}{0.33\linewidth}
        \centering
        \includegraphics[height=28mm]{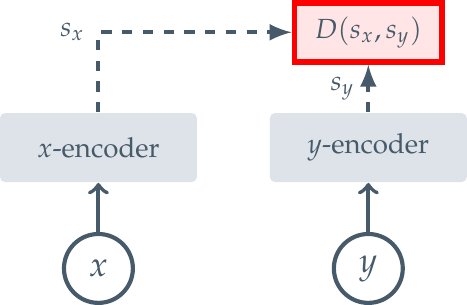}
        \caption{\bf Joint-Embedding Architecture}
        \label{subf-fig:train-loss}
        \label{subfig:jea}
    \end{subfigure}\hfill
    \begin{subfigure}{0.33\linewidth}
        \centering
        \includegraphics[height=28mm]{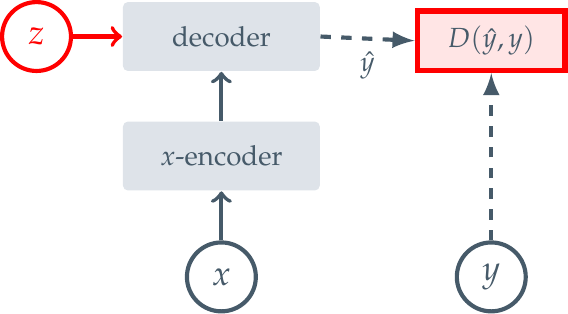}
        \caption{\bf Generative Architecture}
        \label{subfig:generative}
    \end{subfigure}\hfill
    \begin{subfigure}{0.33\linewidth}
        \centering
        \includegraphics[height=28mm]{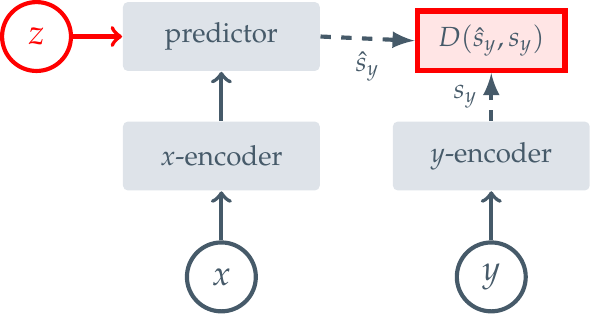}
        \caption{\bf Joint-Embedding Predictive Architecture}
        \label{subfig:jepa}
    \end{subfigure}
    \caption{Common architectures for self-supervised learning, in which the system learns to capture the relationships between its inputs. The objective is to assign a high energy (large scaler value) to incompatible inputs, and to assign a low energy (low scaler value) to compatible inputs. {\bf(a)} Joint-Embedding Architectures learn to output similar embeddings for compatible inputs $x, y$ and dissimilar embeddings for incompatible inputs. {\bf(b)} Generative Architectures learn to directly reconstruct a signal $y$ from a compatible signal $x$, using a decoder network that is conditioned on additional (possibly latent) variables $z$ to facilitate reconstruction. {\bf(c)} Joint-Embedding Predictive Architectures learn to predict the embeddings of a signal $y$ from a compatible signal $x$, using a predictor network that is conditioned on additional (possibly latent) variables $z$ to facilitate prediction.}
    \label{fig:ebm}\vspace{-1em}
\end{figure*}

\section{Introduction}
\label{sec:intro}
In computer vision, there are two common families of approaches for self-supervised learning from images: invariance-based methods~\cite{assran2022masked, he2019moco,caron2020unsupervised,chen2020exploring,grill2020bootstrap,caron2021emerging,zbontar2021barlow,bardes2021vicreg,asano2019self} and generative methods~\cite{devlin2018bert, baevski2022data2vec, pathak2016context, he2021masked}.

Invariance-based pretraining methods optimize an encoder to produce similar embeddings for two or more views of the same image~\cite{bromley1993signature,chen2020simple}, with image views typically constructed using a set of hand-crafted data augmentations, such as random scaling, cropping, and color jittering~\cite{chen2020simple}, amongst others~\cite{grill2020bootstrap}.
These pretraining methods can produce representations of a high semantic level~\cite{caron2021emerging, assran2022masked}, but they also introduce strong biases that may be detrimental for certain downstream tasks or even for pretraining tasks with different data distributions~\cite{assran2022hidden}.
Often, it is unclear how to generalize these biases for tasks requiring different levels of abstraction.
For example, image classification and instance segmentation do not require the same invariances~\cite{bardes2022vicregl}.
Additionally, it is not straightforward to generalize these image-specific augmentations to other modalities such as audio.

Cognitive learning theories have suggested that a driving mechanism behind representation learning in biological systems is
the adaptation of an internal model
to predict sensory input responses~\cite{rao1999predictive,friston2005theory}.
This idea is at the core of self-supervised generative methods, which remove or corrupt portions of the input and learn to predict the corrupted content~\cite{vincent2010stacked,pathak2016context,he2021masked,bao2021beit,xie2021simmim,wei2021masked}.
In particular, mask-denoising approaches learn representations by reconstructing randomly masked patches from an input, either at the pixel or token level.
Masked pretraining tasks require less prior knowledge than view-invariance  approaches and easily generalize beyond the image modality~\cite{baevski2022data2vec}.
However, the resulting representations are typically of a lower semantic level and underperform invariance-based pretraining in off-the-shelf evaluations (e.g., linear-probing) and in transfer settings with limited supervision for semantic classification tasks~\cite{assran2022masked}.
Consequently, a more involved adaptation mechanism (e.g., end-to-end fine-tuning) is required to reap the full advantage of these methods.

In this work, we explore how to improve the semantic level of self-supervised representations without using extra prior knowledge encoded through image transformations. 
To that end, we introduce a joint-embedding predictive architecture~\cite{lecun2022path} for images (\putalg).
An illustration of the method is provided in Figure~\ref{fig:method}.
The idea behind \putalg is to predict missing information in an abstract representation space; e.g., given a single context block, predict the representations of various target blocks in the same image, where target representations are computed by a learned target-encoder network.

Compared to generative methods that predict in pixel/token space, \putalg makes use of abstract prediction targets for which unnecessary pixel-level details are potentially eliminated, thereby leading the model to learn more semantic features.
Another core design choice to guide \putalg towards producing semantic representations is the proposed multi-block masking strategy. Specifically, we demonstrate the importance of predicting sufficiently large target blocks in the image, using an informative (spatially distributed) context block.

Through an extensive empirical evaluation, we demonstrate that:
\begin{itemize}
\item \putalg learns strong off-the-shelf representations without the use of hand-crafted view augmentations (cf.~Fig.\ref{fig:mae-comparison}).
\putalg outperforms pixel-reconstruction methods such as MAE~\cite{he2021masked} on ImageNet-1K linear probing, semi-supervised 1\% ImageNet-1K, and semantic transfer tasks. 

\item \putalg is competitive with view-invariant pretraining approaches on semantic tasks and achieves better performance on low-level visions tasks such as object counting and depth prediction (Sections~\ref{sec:classification} and~\ref{sec:local_prediction}). By using a simpler model with less rigid inductive bias, \putalg is applicable to a wider set of tasks.

\item \putalg is also scalable and efficient (Section~\ref{sec:scale}). Pre-training a ViT-H/14 on ImageNet requires less than 1200 GPU hours, which is over $2.5\times$ faster than a ViT-S/16 pretrained with iBOT\cite{zhou2021ibotyes} and over $10\times$ more efficient than a ViT-H/14 pretrained with MAE. Predicting in representation space significantly reduces the total computation needed for self-supervised pretraining.
\end{itemize}

\section{Background}
\label{sec:background}

Self-supervised learning is an approach to representation learning in which a system learns to capture the relationships between its inputs.
This objective can be readily described using the framework of Energy-Based Models (EBMs)~\cite{lecun2006tutorial} in which the self-supervised objective is to assign a high energy to incompatible inputs, and to assign a low energy to compatible inputs.
Many existing generative and non-generative approaches to self-supervised learning can indeed be cast in this framework; see Figure~\ref{fig:ebm}.

\paragraph{Joint-Embedding Architectures.}
Invariance-based pretraining can be cast in the framework of EBMs using a Joint-Embedding Architecture (JEA), which learns to output similar embeddings for compatible inputs, $\vx,\vy$, and dissimilar embeddings for incompatible inputs; see Figure~\ref{subfig:jea}.
In the context of image-based pretraining, compatible $\vx,\vy$ pairs are typically constructed by randomly applying hand-crafted data augmentations to the same input image~\cite{chen2020simple}.

The main challenge with JEAs is representation collapse, wherein the energy landscape is flat (i.e., the encoder produces a constant output regardless of the input).
During the past few years, several approaches have been investigated to prevent representation collapse, such as contrastive losses that explicitly push apart embeddings of negative examples~\cite{bromley1993signature,he2019moco,chen2020exploring}, non-contrastive losses that minimize the informational redundancy across embeddings~\cite{zbontar2021barlow,bardes2021vicreg}, and clustering-based approaches that maximize the entropy of the average embedding~\cite{caron2021emerging,assran2021semi, assran2022masked}.
There are also heuristic approaches that leverage an asymmetric architectural design between the $x$-encoder and $y$-encoder to avoid  collapse~\cite{chen2020exploring,grill2020bootstrap,baevski2022data2vec}.

\paragraph{Generative Architectures.}
Reconstruction-based methods for self-supervised learning can also be cast in the framework of EBMs using Generative Architectures; see Figure~\ref{subfig:generative}.
Generative Architectures learn to directly reconstruct a signal $\vy$ from a compatible signal $\vx$, using a decoder
network that is conditioned on an additional (possibly latent) variable $\vz$ to facilitate reconstruction.
In the context of image-based pretraining, one common approach in computer vision is to produce compatible $\vx,\vy$ pairs using masking~\cite{he2016deep,bao2021beit} where $\vx$ is a copy of the image $\vy$, but with some of the patches masked.
The conditioning variable $\vz$ then corresponds to a set of (possibly learnable) mask and position tokens, that specifies to the decoder which image patches to reconstruct.
Representation collapse is not a concern with these architectures as long as the informational capacity of $\vz$ is low compared to the signal $\vy$.

\paragraph{Joint-Embedding Predictive Architectures.}
As shown in Figure~\ref{subfig:jepa}, Joint-Embedding Predictive Architectures~\cite{lecun2022path} are conceptually similar to Generative Architectures; however, a key difference is that the loss function is applied in embedding space, not input space.
JEPAs learn to predict the embeddings of a signal $\vy$ from a compatible signal $\vx$, using a predictor network that is conditioned on an additional (possibly latent) variable $\vz$ to facilitate prediction.
Our proposed \putalg provides an instantiation of this architecture in the context of images using masking; see Figure~\ref{fig:method}.

In contrast to Joint-Embedding Architectures, JEPAs do not seek representations invariant to a set of hand-crafted data augmentations, but instead seek representations that are predictive of each other when conditioned on additional information $\vz$.
However, as with Joint-Embedding Architectures, representation collapse is also a concern with JEPAs; we leverage an asymmetric architecture between the $\vx$- and $\vy$-encoders to avoid representation collapse.

\section{Method}
\label{sec:method}

\begin{figure}[t]
    \centering
    \includegraphics[width=\linewidth]{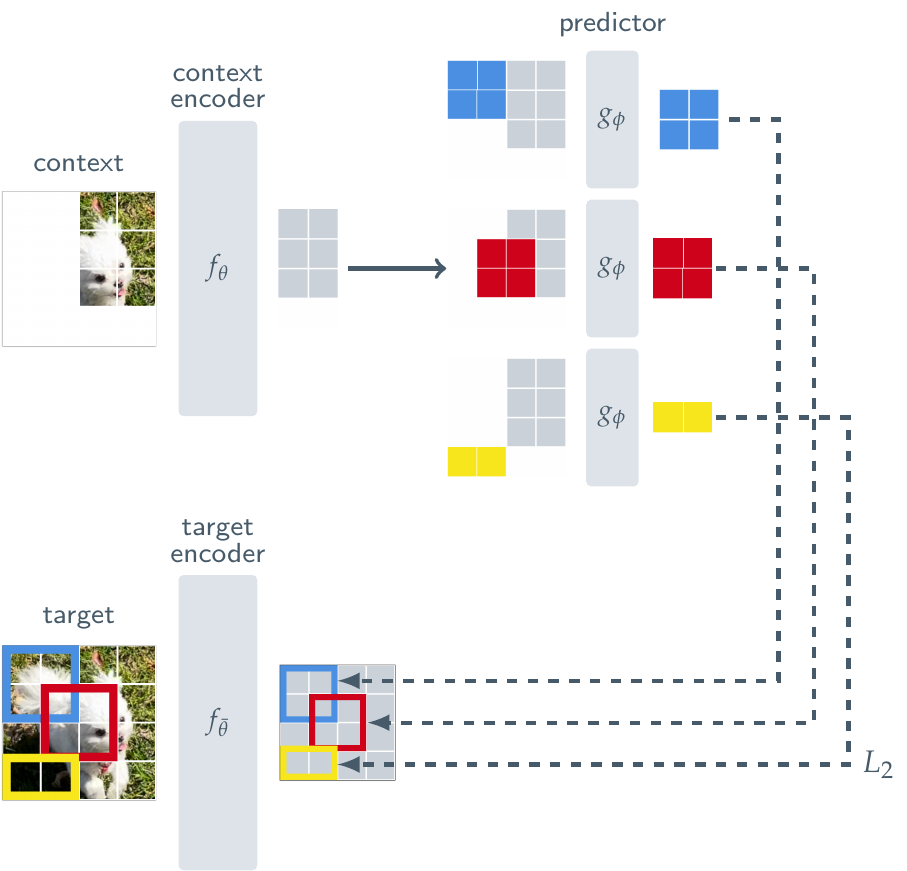}
    \caption{{\bf \putalg.}
    The Image-based Joint-Embedding Predictive Architecture uses a single context block to predict the representations of various target blocks originating from the same image.
    The context encoder is a Vision Transformer (ViT), which only processes the visible context patches.
    The predictor is a narrow ViT that takes the context encoder output and, conditioned on positional tokens (shown in color), predicts the representations of a target block at a specific location.
    The target representations correspond to the outputs of the target-encoder, the weights of which are updated at each iteration via an exponential moving average of the context encoder weights.}
    \label{fig:method}
\end{figure}
We now describe the proposed Image-based Joint-Embedding Predictive Architecture (\putalg), illustrated in Figure~\ref{fig:method}.
The overall objective is as follows: given a context block, predict the representations of various target blocks in the same image.
We use a Vision Transformer~\cite{touvron2021training, dosovitskiy2020image} (ViT) architecture for the context-encoder, target-encoder, and predictor.
A ViT is composed of a stack of transformer layers, each consisting of a self-attention~\cite{vaswani2017attention} operation followed by a fully-connected MLP. 
Our encoder/predictor architecture is reminiscent of the generative masked autoencoders (MAE)~\cite{he2021masked} method. However, one key difference is that the \putalg method is non-generative and the predictions are made in representation space.
\begin{figure}[t]
    \centering
    \includegraphics[width=\linewidth]{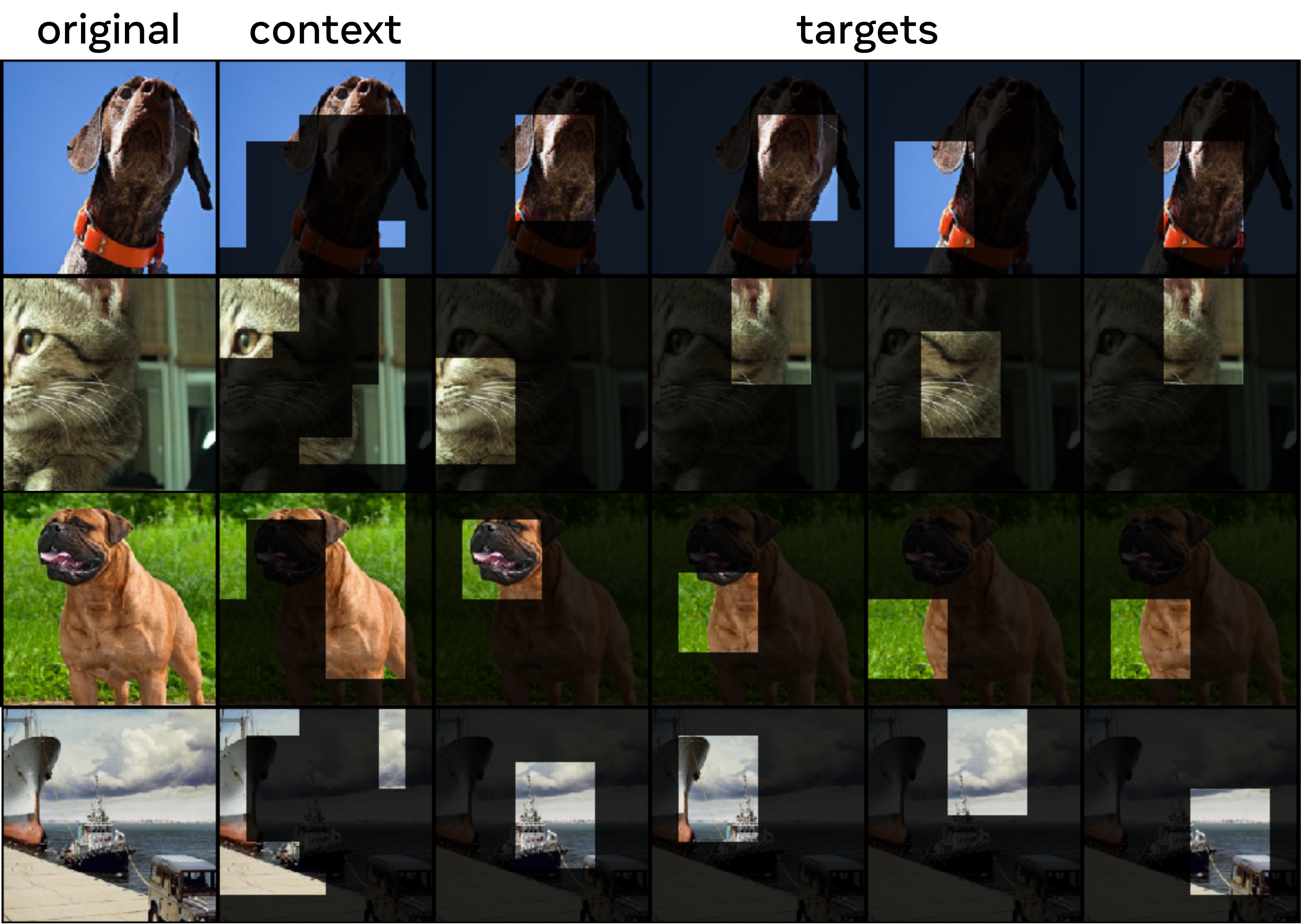}
    \caption{Examples of our context and target-masking strategy. Given an image, we randomly sample $4$ target blocks with scale in the range $(0.15, 0.2)$ and aspect ratio in the range $(0.75, 1.5)$. Next, we randomly sample a context block with scale in the range $(0.85, 1.0)$ and remove any overlapping target blocks. Under this strategy, the target-blocks are relatively semantic, and the context-block is informative, yet sparse (efficient to process).}
    \label{fig:masks}
\end{figure}

\paragraph{Targets.}
We first describe how we produce the targets in the \putalg framework: in \putalg, the targets correspond to the representations of image blocks.
Given an input image $\vy$, we convert it into a sequence of $N$ non-overlapping patches, and feed this through the target-encoder $f_{\bar{\theta}}$ to obtain a corresponding patch-level representation $\vs_y=\{\vs_{y_1}, \dots, \vs_{y_N}\}$ where $\vs_{y_k}$ is the representation associated with the $k^\text{th}$ patch.
To obtain the targets for our loss, we randomly sample $M$ (possibly overlapping) blocks from the target representations $\vs_y$.
We denote by $B_i$ the mask corresponding of the $i^\text{th}$ block and by $\vs_y(i)=\{\vs_{y_{j}}\}_{j \in B_i}$ its patch-level representation.
Typically, we set $M$ equal to 4, and sample the blocks with a random aspect ratio in the range $(0.75, 1.5)$ and random scale in the range $(0.15, 0.2)$.
Note that the target blocks are obtained by masking the \emph{output} of the target-encoder, not the input.
This distinction is crucial to ensure target representations of a high semantic level; see, e.g.,~\cite{baevski2022data2vec}.

\paragraph{Context.}
Recall, the goal behind \putalg is to predict the target block representations from a single context block.
To obtain the context in \putalg, we first sample a single block $\vx$ from the image with a random scale in the range $(0.85, 1.0)$ and unit aspect ratio. We denote by $B_x$ the mask associated with the context block $\vx$.
Since the target blocks are sampled independently from the context block, there may be significant overlap.
To ensure a non-trivial prediction task, we remove any overlapping regions from the context block.
Figure~\ref{fig:masks} shows examples of various context and target blocks in practice.
Next, the masked context block, $\vx$, is fed through the context encoder  $f_\theta$ to obtain a corresponding patch-level representation $\vs_x=\{\vs_{x_j}\}_{j \in B_x}$.

\paragraph{Prediction.}
Given the output of the context encoder, $\vs_x$, we wish to predict the $M$ target block representations $\vs_y(1), \ldots, \vs_y(M)$.
To that end, for a given target block $\vs_y(i)$ corresponding to a target mask $B_i$, the predictor $g_\phi(\cdot, \cdot)$ takes as input the output of the context encoder $\vs_x$ and a  mask token for each patch we wish to predict, $\{\vm_j\}_{j \in B_i}$, and outputs a patch-level prediction $\hat{\vs}_y(i)=\{\hat{\vs}_{y_j}\}_{j \in B_i} = g_\phi(\vs_x, \{\vm_j\}_{j \in B_i})$.
The mask tokens are parameterized by a shared learnable vector with an added positional embedding.
Since we wish to make predictions for $M$ target blocks, we apply our predictor $M$ times, each time conditioning on the mask tokens corresponding to the target-block locations we wish to predict, and obtain predictions $\hat{\vs}_y(1), \ldots, \hat{\vs}_y(M)$.

\paragraph{Loss.}
The loss is simply the average $L_2$ distance between the predicted patch-level representations $\hat{\vs}_y(i)$ and the target patch-level representation $\vs_y(i)$; i.e.,
\[
    \frac{1}{M} \sum^M_{i=1} {D}\left(\hat{\vs}_y(i), \vs_y(i)\right) = \frac{1}{M}  \sum^M_{i=1}\sum_{j \in B_i} \lVert\hat{\vs}_{y_j} - \vs_{y_j}\rVert^2_2.
\]
The parameters of the predictor, $\phi$, and context encoder, $\theta$, are learned through gradient-based optimization, while the parameters of the target encoder $\bar{\theta}$ are updated via an exponential moving average of the context-encoder parameters.
The use of an exponential moving average target-encoder has proven essential for training JEAs with Vision Transformers~\cite{chen2021empirical,zhou2021ibotyes,caron2021emerging}, we find the same to be true for \putalg.

\section{Related Work}
\label{sec:related_works}

A long line of work has explored visual representation learning by predicting the values of missing or corrupted sensory inputs.
Denoising autoencoders use random noise as input corruption~\cite{vincent2010stacked}.
Context encoders regress an entire image region based on its surrounding~\cite{pathak2016context}.
Other works cast image colorization as a denoising task~\cite{zhang2016colorful,larsson2016learning,larsson2017colorization}.

The idea of image denoising has recently been revisited in the context of masked image modelling~\cite{he2021masked, xie2021simmim, bao2021beit}, where a Vision Transformer~\cite{dosovitskiy2020image} is used to reconstruct missing input patches.
The work on Masked Autoencoders (MAE)~\cite{he2021masked} proposed an efficient architecture that only requires the encoder to process visible image patches.
By reconstructing missing patches in pixels space, MAE achieves strong performance when fine-tuned end-to-end on large labeled datasets and exhibits good scaling properties.
 BEiT~\cite{bao2021beit} predicts the value of missing patches in a tokenized space; specifically, tokenizing image patches using a frozen discreteVAE, which is trained on a dataset containing 250 million images ~\cite{ramesh2021zero}.
Yet, pixel-level pre-training has been shown to outperform BEiT for fine-tuning~\cite{he2021masked}.
Another work, SimMIM~\cite{xie2021simmim}, explores reconstruction targets based on the classic Histogram of Gradients~\cite{dalal2005histograms} feature space, and demonstrates some advantage over pixel space reconstruction.
Different from those works, our representation space is learned during training through a Joint-Embedding Predictive Architecture.
Our goal is to learn semantic representations that do not require extensive fine-tuning on downstream tasks.

\begin{table}[t]
    \centering
    \begin{tabular}{l l l c }
        \bf\small Method & \bf\small Arch. & \bf\small Epochs & \bf\small Top-1\\
        \toprule
        \multicolumn{4}{l}{\small\bf\it Methods without view data augmentations}\\
        \small data2vec~\cite{baevski2022data2vec} & \small ViT-L/16 & 1600 & 77.3\\[1ex]
        \multirow{3}{*}{\small MAE~\cite{he2021masked}} & \small ViT-B/16 & 1600 & 68.0\\
        & \small ViT-L/16 & 1600 & 76.0\\
        & \small ViT-H/14 & 1600 & 77.2\\[1ex]
        \multirow{2}{*}{\small CAE~\cite{chen2022context}} & \small ViT-B/16 & 1600 & 70.4\\
         & \small ViT-L/16 & 1600 & 78.1\\[1ex]
        \multirow{3}{*}{\small \putalg} & \cc\small ViT-B/16 & \cc 600 & \cc {72.9}\\
        & \cc\small ViT-L/16 & \cc 600  & \cc{77.5}\\
        & \cc\small ViT-H/14 &  \cc 300 & \cc{79.3}\\
        & \cc\small ViT-H/$16_{448}$ &  \cc 300 & \cc{\bf 81.1}\\
        \midrule
        \multicolumn{4}{l}{\small\bf\it Methods using extra view data augmentations}\\
        \small SimCLR v2~\cite{chen2020big} & \small RN152 ($2\times$) & 800 & 79.1 \\
        \small DINO~\cite{caron2021emerging} & \small ViT-B/8 & 300 & 80.1 \\
        \small iBOT~\cite{zhou2021ibotyes} & \small ViT-L/16 & 250 & \bf 81.0
    \end{tabular}
    \caption{{\bf ImageNet}. Linear-evaluation on ImageNet-1k (the ViT-H/16$_{448}$ is pretrained at at a resolution of $448 \times 448$). \putalg improves linear probing performance compared to other methods that do not rely on hand-crafted view data-augmentations during pretraining. Moreover, \putalg demonstrates good scalability --- the larger \putalg model matches the performance of view-invariance approaches without requiring view data-augmentations.}
  \label{tb:lineareval}
\end{table}

Closest to our work is data2vec~\cite{baevski2022data2vec} and Context Autoencoders~\cite{chen2021empirical}.
The data2vec method learns to predict the representation of missing patches computed through an online target encoder; by avoiding handcrafted augmentations, the method can be applied to diverse modalities with promising results in vision, text and speech.
Context Autoencoders use an encoder/decoder architecture optimized via the sum of a reconstruction loss and an alignment constraint, which enforces predictability of missing patches in representation space.
Compared to these methods, \putalg exhibits significant improvements in computational efficiency and learns more semantic off-the-shelf representations.
Concurrent to our work, data2vec-v2~\cite{baevski2022efficient} explores efficient architectures for learning with various modalities.

We also compare \putalg with various methods based on joint-embedding architectures; e.g., DINO~\cite{caron2021emerging}, MSN~\cite{assran2022masked} and iBOT~\cite{zhou2021ibotyes}.
Theses methods rely on hand-crafted data augmentations during pretraining to learn semantic image representations.
The work on MSN~\cite{assran2022masked}, uses masking as an additional data-augmentation during pretraining, while iBOT combines a data2vec-style patch-level reconstruction loss with the DINO view-invariance loss.
Common to these approaches is the need to process multiple user-generated views of each input image, thereby hindering scalability.
By contrast, \putalg only requires processing a single view of each image.
We find that a ViT-Huge/14 trained with \putalg requires less computational effort than a ViT-Small/16 trained with iBOT.

\section{Image Classification}
\label{sec:classification}

\begin{table}[t]
    \centering
    \begin{tabular}{l l l c}
        \bf\small Method & \bf\small Arch. & \bf\small Epochs & \bf\small Top-1\\
        \toprule
        \multicolumn{4}{l}{\small\bf\it Methods without view data augmentations}\\
        \small data2vec~\cite{baevski2022data2vec} & \small ViT-L/16 & 1600& 73.3\\[1ex]
        \multirow{2}{*}{\small MAE~\cite{he2021masked}} & \small ViT-L/16 & 1600 & 67.1\\
        & \small ViT-H/14 & 1600 & 71.5\\[1ex]
        \multirow{3}{*}{\small \putalg} & \cc\small ViT-L/16 & \cc 600 & \cc 69.4\\
        & \cc\small ViT-H/14 & \cc 300 & \cc  73.3\\
        & \cc\small ViT-H/$16_{448}$ & \cc 300 & \cc \bf 77.3\\
        \midrule
        \multicolumn{4}{l}{\small\bf\it Methods using extra view data augmentations}\\
        \small iBOT~\cite{zhou2021ibotyes} & \small ViT-B/16 & 400 &  69.7\\
        \small DINO~\cite{caron2021emerging} & \small ViT-B/8 & 300 &  70.0 \\
        \small SimCLR v2~\cite{grill2020bootstrap} & \small RN151 ($2\times$) & 800 & 70.2 \\
        \small BYOL~\cite{grill2020bootstrap} & \small RN200 ($2\times$) & 800 & 71.2 \\
        \small MSN~\cite{assran2022masked} & \small ViT-B/4 & 300 &  \bf 75.7
    \end{tabular}
    \caption{{\bf ImageNet-1\%}. Semi-supervised evaluation on ImageNet-1K using only 1\% of the available labels. Models are adapted via fine-tuning or linear-probing, depending on whichever works best for each respective method. ViT-H/16$_{448}$ is pretrained at at a resolution of $448\times 448$. \putalg pretraining outperforms MAE which also does not rely on hand-crafted data-augmentations during pretraining. Moreover, \putalg benefits from scale. A ViT-H/16 trained at resolution $448$ surpasses previous methods including methods that leverage extra hand-crafted data-augmentations.}
    \label{tb:lowshot}
\end{table}

To demonstrate that \putalg learns high-level representations without relying on hand-crafted data-augmentations, we report results on various image classification tasks using the linear probing and partial fine-tuning protocols.
In this section, we consider self-supervised models that have been pretrained on the ImageNet-1K dataset~\cite{russakovsky2015imagenet}.
Pretraining and evaluation implementation details are described in the Appendix~\ref{app:implementation_details}.
All \putalg models are trained at resolution $224 \times 224$ pixels, unless stated otherwise.

\paragraph{ImageNet-1K.}
Table~\ref{tb:lineareval} shows performance on the common ImageNet-1K linear-evaluation benchmark.
After self-supervised pretraining, the model weights are frozen and a linear classifier is trained on top using the full ImageNet-1K training set.
Compared to popular methods such as Masked Autoencoders (MAE)~\cite{he2021masked}, Context Autoencoders (CAE)~\cite{chen2022context}, and data2vec~\cite{baevski2022data2vec}, which also do not rely on extensive hand-crafted data-augmentations during pretraining, we see that \putalg significantly improves linear probing performance, while using less computational effort (see section~\ref{sec:scale}).
By leveraging the improved efficiency of \putalg, we can train larger models that outperform the best CAE model while using a fraction of the compute.
\putalg also benefits from scale; in particular, a ViT-H/16 trained at resolution $448 \times 448$ pixels matches the performance of view-invariant approaches such as iBOT~\cite{zhou2021ibotyes}, despite avoiding the use of hand-crafted data-augmentations.

\paragraph{Low-Shot ImageNet-1K.}
Table~\ref{tb:lowshot} shows performance on the 1\% ImageNet benchmark.
Here the idea is to adapt the pretrained models for ImageNet classification using only 1\% of the available ImageNet labels, corresponding to roughly 12 or 13 images per class.
Models are adapted via fine-tuning or linear-probing, depending on whichever works best for each respective method.
\putalg outperforms MAE while requiring less pretraining epochs when using a similar encoder architecture.
\putalg, using a ViT-H/14 architecture,  matches the performance of a ViT-L/16 pretrained with data2vec~\cite{baevski2022data2vec}, while using significantly less computational effort (see Section~\ref{sec:scale}).
By increasing the image input resolution, \putalg outperforms previous methods including joint-embedding methods that do leverage extra hand-crafted data-augmentations during pretraining, such as MSN~\cite{assran2022masked}, DINO~\cite{caron2020unsupervised}, and iBOT~\cite{zhou2021ibotyes}.

\paragraph{Transfer learning.}
Table~\ref{tb:transfer-classification} shows performance on various downstream image classification tasks using a linear probe.
\putalg significantly outperforms previous methods that do not use augmentations (MAE and data2vec), and decreases the gap with the best view-invariance-based methods, which leverage hand-crafted data augmentations during pretraining, even surpassing the popular DINO~\cite{caron2021emerging} on CIFAR100 and Place205 with a linear probe.
\begin{table}[t]
    \centering
    \begin{tabular}{l l c c c}
        \bf\small Method & \bf\small Arch. & \bf\small CIFAR100 & \bf\small Places205 &  \bf\small iNat18 \\
        \toprule
        \multicolumn{5}{l}{\small\bf\it Methods without view data augmentations}\\
        \small data2vec~\cite{baevski2022data2vec} & \small ViT-L/16 & 81.6 & 54.6 & 28.1 \\
        \small MAE~\cite{he2021masked} & \small ViT-H/14 & 77.3 & 55.0 & 32.9 \\
        \small \putalg & \cc\small ViT-H/14 & \cc\bf 87.5 & \cc\bf 58.4 & \cc\bf 47.6 \\
        \midrule
        \multicolumn{4}{l}{\small\bf\it Methods using extra view data augmentations}\\
        \small DINO~\cite{caron2021emerging} & \small ViT-B/8 & 84.9 & 57.9 & 55.9 \\
        \small iBOT~\cite{zhou2021ibotyes} & \small ViT-L/16 & \bf 88.3 & \bf 60.4 & \bf 57.3
    \end{tabular}
    \caption{\textbf{Linear-probe transfer for image classification}. Linear-evaluation on downstream image classification tasks. \putalg significantly outperforms previous methods that also do not use augmentations (MAE and data2vec), and decreases the gap with the best view-invariance-based methods that leverage hand-crafted data augmentations during pretraining.}
    \label{tb:transfer-classification}
\end{table}

\section{Local Prediction Tasks}
\label{sec:local_prediction}
As demonstrated in Section~\ref{sec:classification}, \putalg learns semantic image representations that significantly improve the downstream image classification performance of previous methods, such as MAE and data2vec. Additionally, \putalg benefits from scale and can close the gap, and even surpass, view-invariance based methods that leverage extra hand-crafted data augmentations.
In this section, we find that \putalg also learns local image features and surpasses view-invariance based methods on low-level and dense prediction tasks, such as object counting and depth prediction.

Table~\ref{tb:transfer-lowlevel} shows performance on various low-level tasks using a linear probe.
After pretraining, the encoder weights are frozen and a linear model is trained on top to perform object-counting and depth prediction on the Clevr dataset~\cite{clevr}.
Compared to view-invariance methods such as DINO and iBOT, the \putalg method effectively captures low-level image features during pretraining and outperforms them in object counting (Clevr/Count) and (by a large margin) depth prediction (Clevr/Dist).
\begin{table}[t]
    \centering
    \begin{tabular}{l l c c}
        \bf\small Method & \bf\small Arch. & \bf\small Clevr/Count & \bf\small Clevr/Dist \\
        \toprule
        \multicolumn{4}{l}{\small\bf\it Methods without view data augmentations}\\
        \fcg \small data2vec~\cite{baevski2022data2vec} & \fcg \small ViT-L/16 & \fcg 85.3 & \fcg 71.3 \\
        \fcg \small MAE~\cite{he2021masked} & \fcg \small ViT-H/14 & \fcg\bf  90.5 & \fcg \bf 72.4 \\
        \small \putalg & \cc\small ViT-H/14 & \cc 86.7 & \cc\bf 72.4 \\
        \midrule
        \multicolumn{4}{l}{\small\bf\it Methods using extra data augmentations}\\
        \small DINO~\cite{caron2021emerging} & \small ViT-B/8 & 86.6 & 53.4 \\
        \small iBOT~\cite{zhou2021ibotyes} & \small ViT-L/16 & 85.7 & 62.8
    \end{tabular}
    \caption{\textbf{Linear-probe transfer for low-level tasks}. Linear-evaluation on downstream low-level tasks consisting of object counting (Clevr/Count) and depth prediction (Clevr/Dist). The \putalg method effectively captures low-level image features during pretraining and outperforms view-invariance based methods on tasks such object counting and depth prediction.}
    \label{tb:transfer-lowlevel}
\end{table}

\begin{figure}[t]
    \centering
    \includegraphics[width=\linewidth]{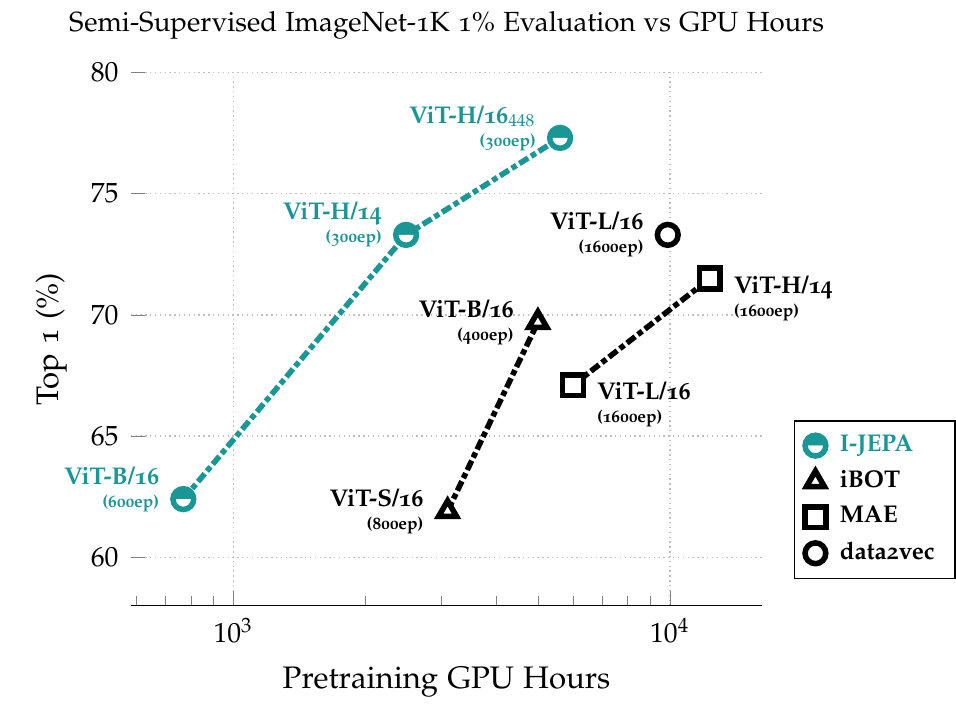}
    \caption{{\bf Scaling}.
    Semi-supervised evaluation on ImageNet-1K 1\% as a function of pretraining GPU hours.
    \putalg requires less compute than previous methods to achieve strong performance.
    Compared to MAE and data2vec, \putalg obtains a significant speedup by requiring fewer pretraining epochs.
    Compared to iBOT, which relies on hand-crafted data-augmentations, a huge \putalg model (ViT-H/14) requires less compute than their smallest model (ViT-S/16).}
    \label{fig:scaling}
\end{figure}
\begin{table*}[h]
\centering
\begin{tabular}{l l c c c c c c}
    \bf\small Pretrain & \bf\small Arch. &  \bf\small CIFAR100 & \bf\small Place205 & \bf\small INat18 & \bf\small Clevr/Count & \bf\small Clevr/Dist\\
    \toprule
    \small IN1k & \small ViT-H/14 & 87.5 & 58.4 & 47.6 & 86.7 & 72.4 \\
    \small IN22k & \small ViT-H/14 & \bf 89.5 & 57.8 & 50.5 & \bf 88.6 & \bf 75.0\\
    \small IN22k & \small ViT-G/16 & \bf 89.5 & \bf 59.1 & \bf 55.3 & 86.7 & 73.0\\
\end{tabular}
\caption{\textbf{Ablating dataset and model size}. Evaluating impact of pre-training dataset size and model size on transfer tasks. \putalg benefits from larger more diverse datasets. When increasing the size of the pretraining dataset (IN1k versus IN22k) we see an performance improvement for the ViT-H/14 model. We observe a further performance improvement on semantic tasks by training a larger model ViT-G/16 model on ImageNet-22k. The ViT-H/14 is trained for 300 epochs on IN1k and the equivalent of 900 IN1K epochs on IN22k. The ViT-H/16 is trained for the equivalent of 600 IN1k epochs.}
\label{tb:transfer-in22k}
\end{table*}

\begin{figure*}[t]
\centering
\includegraphics[width=\linewidth]{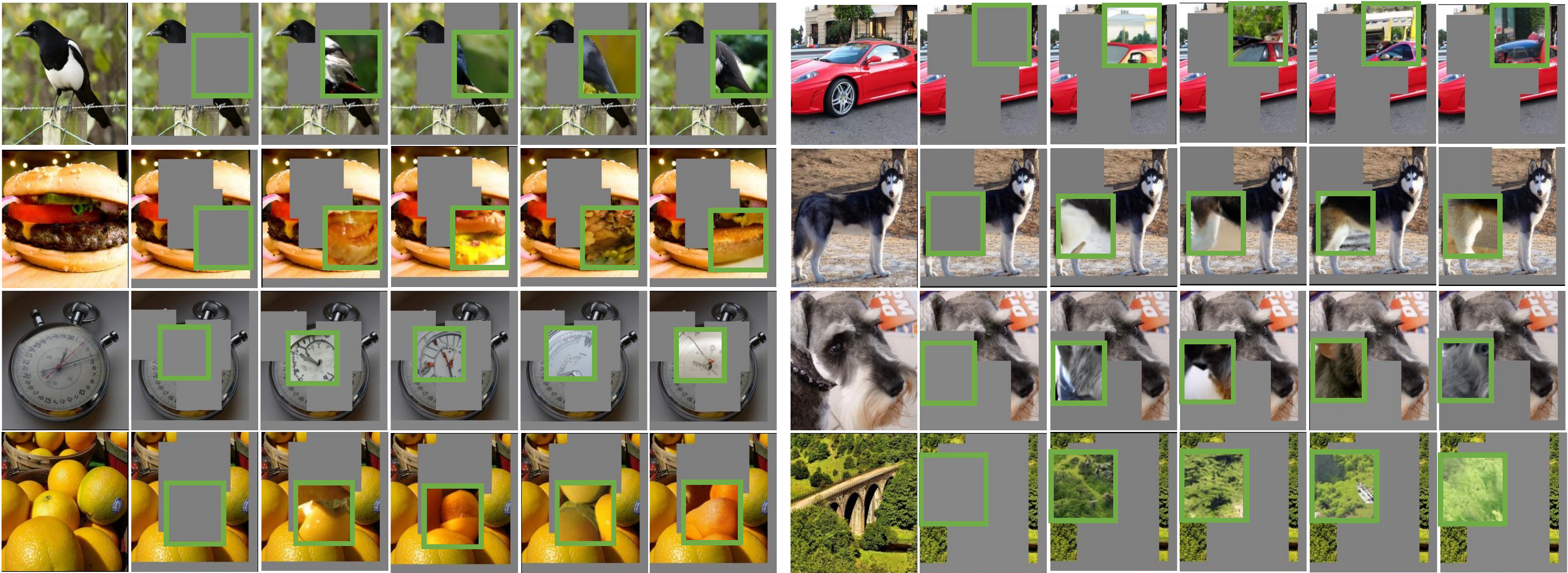}
\caption{\textbf{Visualization of \putalg predictor representations}.
    For each image: first column contains the original image; second column contains the context image, which is processed by a pretrained \putalg ViT-H/14 encoder.
    Green bounding boxes in subsequent columns contain samples from a generative model decoding the output of the pretrained \putalg predictor, which is conditioned on positional mask tokens corresponding to the location of the green bounding box.
    Qualities that are common across samples represent information that contained is in the \putalg prediction.
    The \putalg predictor is correctly capturing positional uncertainty and producing high-level object parts with a correct pose (e.g., the back of the bird and top of a car).
    Qualities that vary across samples represent information that is not contained in the representation.
    In this case, the \putalg predictor discards the precise low-level details as well as background information.} 
\label{fig:visualization-predictor}
\end{figure*}

\section{Scalability}
\label{sec:scale}

\paragraph{Model Efficiency.}
\putalg is highly scalable compared to previous approaches.
Figure~\ref{fig:scaling} shows semi-supervised evaluation on 1\% ImageNet-1K as a function of GPU hours.
\putalg requires less compute than previous methods and achieves strong performance without relying on hand-crafted data-augmentations.
Compared to reconstruction-based methods, such as MAE, which directly use pixels as targets, \putalg introduces extra overhead by computing targets in representation space (about 7\% slower time per iteration).
However, since \putalg converges in roughly $5\times$ fewer iterations, we still see significant compute savings in practice.
Compared to view-invariance based methods, such as iBOT, which rely on hand-crafted data augmentations to create and process multiple views of each image, \putalg also runs significantly faster.
In particular, a huge \putalg model (ViT-H/14) requires less compute than a small iBOT model (ViT-S/16).

\paragraph{Scaling data size.}
We also find \putalg to benefit from pretraining with larger datasets.
Table~\ref{tb:transfer-in22k} shows transfer learning performance on semantic  and low level tasks when increasing the size of the pretraining dataset (IN1K versus IN22K).
Transfer learning performance on these conceptually different tasks improves when pretraining on a larger more diverse dataset.

\paragraph{Scaling model size.}
Table~\ref{tb:transfer-in22k} also shows that \putalg benefit from larger model size when pretraining on IN22K. Pretraining a ViT-G/16 significantly improves the downstream performances on image classification tasks such as Place205 and INat18 compared to a ViT-H/14 model, but does not improve performance on low-level downstream tasks --- the ViT-G/16 uses larger input patches, which can be detrimental for the local prediction tasks.

\begin{table*}[h]
    \centering
    \begin{tabular}{l l c l c c}
        & \multicolumn{2}{l}{\bf\small Targets} & \multicolumn{2}{l}{\bf\small Context} \\ \cmidrule(lr){2-3}\cmidrule(lr){4-5}
        \bf\small Mask & \small Type & \small Freq. & \small Type & \small Avg. Ratio$^*$ & \bf\small Top-1 \\
        \toprule
        \cc\tt multi-block & \cc\small Block$(0.15, 0.2)$ & \cc\small 4 & \cc\small Block$(0.85,1.0)$ $\times$ Complement & \cc 0.25 & \cc\bf 54.2 \\
        \small\tt rasterized & \small Quadrant & \small 3 & \small Complement & 0.25 & 15.5 \\
        \small\tt block & \small Block$(0.6)$ & \small 1 & \small Complement & 0.4 & 20.2 \\
        \small\tt random & \small Random(0.6) & \small 1 & \small Complement & 0.4 & 17.6 \\
        \bottomrule
        \multicolumn{6}{c}{\footnotesize \it $^*$Avg. Ratio is the average number of patches in the context block relative to the total number of patches in the image.}
    \end{tabular}
    \caption{\textbf{Ablating masking strategy}.
    Linear evaluation on ImageNet-1K using only 1\% of the available labels
    after \putalg pretraining of a ViT-B/16 for 300 epochs. 
    Comparison of proposed {\tt multi-block} masking strategy. In {\tt rasterized} masking the image is split into four large quadrants; one quadrant is used as a context to predict the other three quadrants.
    In {\tt block} masking, the target is a single image block and the context is the image complement.
    In {\tt random} masking, the target is a set of random image patches and the context is the image complement.
    The proposed {\tt multi-block} masking strategy is helpful for guiding I-JEPA to learn semantic representations.}
    \label{tb:ablation-masking}
\end{table*}

\section{Predictor Visualizations}
\label{sec:predictor}

The role of the predictor in \putalg is to take the output of the context encoder and, conditioned on positional mask tokens, to predict the representations of a target black at the location specified by the mask tokens.
One natural question is whether the predictor conditioned on the positional mask tokens is learning to correctly capture positional uncertainty in the target.
To qualitatively investigate this question, we visualize the outputs of the predictor.
We use the following visualization approach to enable the research community to independently reproduce our findings.
After pretraining, we freeze the context-encoder and predictor weights, and train a decoder following the RCDM framework~\cite{bordes2022high} to map the average-pool of the predictor outputs back to pixel space.
Figure~\ref{fig:visualization-predictor} shows decoder outputs for various random seeds.
Qualities that are common across samples represent information that is contained in the average-pooled predictor representation.
The \putalg predictor correctly captures positional uncertainty and produces high-level object parts with the correct pose (e.g., back of the bird and top of the car).

\section{Ablations}
\label{sec:ablations}

\begin{table}[!h]
    \centering
    \begin{tabular}{l l c c}
        \bf\small Targets & \bf\small Arch. & \bf\small Epochs & \bf\small Top-1 \\
        \toprule
        \cc \small Target-Encoder Output & \cc\small ViT-L/16 & \cc 500 & \cc\bf 66.9 \\
        \small Pixels & \small ViT-L/16 & 800 & 40.7
    \end{tabular}
    \caption{\textbf{Ablating targets}.
    Linear evaluation on ImageNet-1K using only 1\% of the available labels.
    The semantic level of the \putalg representations degrades significantly when the loss is applied in pixel space, rather than representation space, highlighting the importance of the target-encoder during pretraining.}
    \label{tb:ablation-rep-space}
\end{table}
\paragraph{Predicting in representation space.}
Table~\ref{tb:ablation-rep-space} compares low-shot performance on 1\% ImageNet-1K using a linear probe when the loss is computed in pixel-space versus representation space.
We conjecture that a crucial component of \putalg is that the loss is computed entirely in representation space, thereby giving the target encoder the ability to produce abstract prediction targets, for which irrelevant pixel-level details are eliminated.
From Table~\ref{tb:ablation-rep-space}, it is clear that predicting in pixel-space leads to a significant degradation in the linear probing performance.

\paragraph{Masking strategy.}

Table~\ref{tb:ablation-masking} compare our {\tt multi-block} masking with other masking strategies
such as {\tt rasterized} masking, where the image is split into four large quadrants, and the goal is to use one quadrant as a context to predict the other three quadrants,
and the  traditional {\tt block} and {\tt random} masking typically used in reconstruction-based methods.
In {\tt block} masking, the target is a single image block and the context is the image complement.
In {\tt random} masking, the target is a set of random patches and the context is the image complement.
Note that there is no overlap between the context and target blocks in all considered strategies.
We find {\tt multi-block} masking helpful for guiding \putalg to learning semantic representations.
Additional ablations on {\tt multi-block} masking can be found in Appendix~\ref{appdn:ablation}.

\section{Conclusion}
\label{sec:conclusion}
We proposed \putalg, a simple and efficient method for learning semantic image representations without relying on hand-crafted data augmentations.
We show that by predicting in representation space, \putalg converges faster than pixel reconstruction methods and learns representations of a high semantic level.
In contrast to view-invariance based methods, \putalg highlights a path for learning general representations with joint-embedding architectures, without relying on hand-crafted view augmentations.

{\small
\bibliographystyle{ieee_fullname}
\bibliography{egbib}}
\vfill
\clearpage
\onecolumn

\appendix

\section{Implementation Details}
\label{app:implementation_details}

\subsection{Pretraining}

\paragraph{Architectures.}
For \putalg pretraining, we use Vision Transformer~\cite{dosovitskiy2020image} (ViT) architectures for the context-encoder, target-encoder, and the predictor.
While the context-encoders and target-encoders correspond to standard ViT architectures, the predictor is designed as a light-weight (narrow) ViT architecture.
Specifically, we fix the embedding dimension of the predictor to 384, while keeping the number of self-attention heads equal to that of the backbone context-encoder.
For the smaller ViT-B/16 context-encoder, we set the depth of the predictor to 6.
For ViT-L/16, ViT-H/16, and ViT-H/14 context-encoders, we set the depth of the predictor to 12. Finally, the ViT-G/16 uses a predictor of depth 16.
\putalg is pretrained without a {\tt [cls]} token. We use the target-encoder for evaluation and average pool its output to produce a global image representation.

\paragraph{Optimization.}
We use AdamW~\cite{loshchilov2017decoupled} to optimize the context-encoder and predictor weights.
Our default batch-size is 2048, and the learning rate is linearly increased from $10^{-4}$ to $10^{-3}$ during the first 15 epochs of pretraining, and decayed to $10^{-6}$ following a cosine schedule thereafter.
Following~\cite{caron2021emerging, assran2022masked}, the weight-decay is linearly increased from $0.04$ to $0.4$ throughout pretraining.
The target-encoder weights are identical to the context-encoder weights at initialization, and updated via an exponential moving average thereafter~\cite{tarvainen2017mean,he2019moco,chen2020mocov2,grill2020bootstrap,caron2021emerging,assran2022masked}.
We use a momentum value of $0.996$, and linearly increase this value to $1.0$ throughout pretraining, following~\cite{caron2021emerging, assran2022masked}.

\paragraph{Masking.}
By default, we sample $4$ possibly overlapping target blocks masks with random scale in the range $(015, 0.2)$ and aspect ratio in the range $(0.75, 1.5)$.
We sample $1$ context block mask with random scale in the range $(0.85, 1.0)$ and unit aspect ratio.
We subsequently eliminate any regions in the context block mask that overlap with any of the $4$ target block masks.
The context-block mask and target-block masks are sampled independently for each image in the mini-batch.
To ensure efficient batch processing, we restrict the size of all context masks on a co-located GPU to be identical.
Similarly, we restrict the size of all target masks on a co-located GPU to be identical.
The mask-sampler is efficiently implemented in only a few lines of code in PyTorch~\cite{paszke2019pytorch} using a batch-collator function, which runs in the data loader processes.
In short, in each iteration, the data loader returns a mini-batch of images and a set of context and target masks for each image, identifying the patch indices to keep for the context and target views.

\subsection{Downstream Tasks}

\paragraph{Linear evaluation.}
When evaluating methods such as iBOT~\cite{zhou2021ibotyes}, DINO~\cite{caron2021emerging} or MAE~\cite{he2021masked}, which leverage Vision Transformers~\cite{dosovitskiy2020image} with an additional {\tt [cls]} token, we use the default configurations of VISSL~\cite{goyal2021vissl} to evaluate all the models on iNaturalist18~\cite{van2018inaturalist}, CIFAR100~\cite{krizhevsky2009learning}, Clevr/Count~\cite{johnson2017clevr, vtab}, Clevr/Dist~\cite{johnson2017clevr, vtab}, and Places205~\cite{zhou2014learning}.
We freeze the encoder and return the best number among the following representations: 1) the {\tt [cls]} token representation of the last layer, 2) the concatenation of the last $4$ layers of the {\tt [cls]} token.
For each representation, we try two different heads: 1) a linear head, or 2) a linear head preceded by a batch normalization, and return the best number.
We use the default data augmentations of VISSL~\cite{goyal2021vissl}: random resize cropping and horizontal flipping, with the exception of Clevr/Count and Clevr/Dist, where we only use center crop and horizontal flipping, as random cropping interferes with the capability of counting objects and estimating distance, removing critical objects from the scene.
For CIFAR100, we resize the images to $224 \times 224$ pixels, so as to keep the number of patches equal to that used during pretraining.

Because our \putalg implementation uses Vision Transformer architectures without a {\tt [cls]} token, we adapt the default VISSL evaluation recipe to utilize the average-pooled patch representation instead of the {\tt [cls]} token. 
We therefore report the best linear evaluation number among the following representations: 1) the average-pooled patch representation of the last layer, 2) the concatenation of the last $4$ layers of the average-pooled patch representations.
We otherwise keep the linear-probing recipe identical.

\paragraph{ImageNet evaluations.}
To evaluate the \putalg on ImageNet~\cite{russakovsky2015imagenet}, we adapt the VISSL recipe to use average pooled representations instead of the {\tt [cls]} token.
Following MAE~\cite{he2021masked}, we use the LARS~\cite{lars} optimizer with a batch-size of $16384$, and train the linear probe for 50 epochs.
We use a learning rate with a step-wise decay, dividing it by a factor of $10$ every $15$ epochs, and sweep three different reference learning rates $[0.01, 0.05, 0.001]$, and two weight decay values $[0.0005, 0.0]$. 

\paragraph{Low-shot evaluation.} To evaluate our model on the ImageNet-1\% low-shot task, we adapt the fine-tuning protocol of MAE~\cite{he2021masked}.We fine-tune our ViT-L/H models for 50 epochs on ImageNet-1\% with the AdamW optimizer and a cosine learning rate scheduler. We use a batch size of 512, a learning rate layer decay of $0.75$ and $0.1$ label smoothing. We use the default randaugment data-augmentations as in MAE. In contrast to the fine-tuning done with MAE, we do not use mixup, cutmix, random erasing  or drop path. For the \putalg, we use a learning rate /weight decay of 3e$^{-5}$/5e$^{-2}$ for the ViT-L/16, 3e$^{-5}$/4e$^{-1}$ for the ViT-H/14 and 3e$^{-5}$/4e$^{-1}$ for the ViT-H/16$_{448}$.  Similar fine-tuning strategy for low-shot learning has been explored by Semi-VIT in the context of semi-supervised learning~\cite{cai2022semi}.

\section{Broader Related Work}

Self-supervised learning of visual representations with joint-embedding architectures is an active line of research~\cite{wu2018unsupervised,he2019moco,chen2020exploring,grill2020bootstrap,chen2020mocov2,caron2021emerging,bardes2021vicreg,zhou2021ibotyes,bordes2022guillotine,mitrovic2020representation,assran2020supervision}.
These approaches train a pair of encoders to output similar embeddings for two or more views of the same image.
To avoid pathological solution, many popular joint-embedding approaches use explicit regularization~\cite{chen2020simple, caron2021emerging, bardes2021vicreg, assran2021semi} or architectural constraints~\cite{grill2020bootstrap,chen2020exploring}.
Collapse-prevention based on architectural constraints leverage specific network design choices to avoid collapse, for example, by stopping the gradient flow in one of the joint-embedding branches~\cite{chen2020simple}, using a momentum encoder in one of the joint-embedding branches~\cite{grill2020bootstrap}, or using an asymmetric prediction head~\cite{grill2020bootstrap, chen2020simple, baevski2022data2vec}.
Recent work~\cite{tian2021understanding} attempts to theoretically understand (in certain simplified settings) how joint-embedding methods with architectural constraints avoid representation collapse without explicit regularization.

Typical regularization-based approaches to collapse prevention in joint-embedding architectures try to maximize the volume of space occupied by the representations.
This is often motivated through the InfoMax~\cite{ma2022principles} principle.
Indeed, a longstanding conviction in unsupervised representation learning is that the resulting representations should be both maximally informative about the inputs, while also satisfying certain simplicity constraints~\cite{linsker1988self, goodfellow2016deep}.
The former objective is often referred to as the information-maximization principle (InfoMax), while the latter is sometimes referred to as the parsimony principle~\cite{ma2022principles}.
Such approaches to representation learning have been proposed for decades (e.g.,~\cite{bridle1991unsupervised}), where, historically, simplicity constraints were enforced by encouraging the learned representations to be sparse, low-dimensional, or disentangled, i.e., the individual dimensions of the representation vector should be statistically independent~\cite{goodfellow2016deep}.
Modern approaches enforce the simplicity constraints coupled with InfoMax regularization through self-supervised loss terms~\cite{hjelm2018learning, tschannen2019mutual, bachman2019learning, krause2010discriminative, hu2017learning, oord2018representation}.
One example is the widespread view-invariance penalty~\cite{misra2020self}, often coupled with with independence~\cite{zbontar2021barlow, bardes2021vicreg} or low-dimensionality constraints, e.g., by projecting representations on the unit hypersphere~\cite{chen2020simple,he2019moco,grill2020bootstrap}.
However, despite its proliferation, there have also been many criticisms of the InfoMax principle, especially since it is does not discriminate between different types of information (e.g, noise and semantics)~\cite{assran2022hidden}.
Indeed, the sets of features we wish the model to capture are not always those with the highest marginal entropy (maximal information content).

Orthogonal to the contributions of invariance-based pretraining, another line of work attempts to learn representations by artificially masking parts of the input and training a network to reconstruct the hidden content~\cite{vincent2010stacked}.
Autoregressive models, and denoising autoencoders in particular, predict clean visual inputs from noisy views~\cite{chen2020generative,vincent2010stacked,he2021masked,bao2021beit,baevski2022data2vec}.
Typically, the goal is to predict missing inputs at a pixel level~\cite{dosovitskiy2020image,he2021masked,xie2019unsupervised}, or at a patch token-level, using a tokenizer~\cite{bao2021beit,wei2021masked}.
While these works demonstrate impressive scalability, they usually learn features at a low-level of semantic abstraction compared to joint-embedding approaches~\cite{assran2022masked}.

More recently, a set of approaches attempt to combine both joint-embedding architectures and reconstruction based approaches~\cite{el2021large}, wherein they combine an invariance pretraining loss with a patch-level reconstruction loss, as in the iBOT method~\cite{zhou2021ibotyes}.
Since view-invariance based approaches are typically biased towards learning global image representations, thereby limiting their applicability to other computer vision tasks, the idea is that adding local loss terms can improve performance on other popular tasks in computer vision~\cite{chen2022intra,gidaris2020learning,bardes2022vicregl}.
The framework of contrastive predictive coding~\cite{oord2018representation} is also closely related to this line of work on local loss terms.
In the context of images~\cite{henaff2020data}, here the idea is to use a contrastive objective combined with a convolutional network to discriminate between overlapping image patch representations.
Specifically, the goal is to encourage the representations of an image patch to be predictable of the image patches directly below it, while pushing away the representations of other patch views.
In contrast to that work, the proposed \putalg method is non-contrastive and does not seek to discriminate between image patches.
Rather, the goal is to predict the representations of various target blocks from a single context block.
This is achieved with a Joint-Embedding Predictive Architecture, using a predictor network that is conditioned on positional embeddings corresponding to the location of the target block in the image.
Qualitative experiments in Section~\ref{sec:predictor} show that the predictor network in our architecture learns to correctly perform this local-to-local region feature mapping, and learns to correctly capture positional uncertainty in the image.

\section{Additional Ablations}
\label{appdn:ablation}
This section follows the same experimental protocol as Section~\ref{sec:ablations}.
We report the result of a linear probe with a frozen backbone, trained on the low-shot 1\% ImageNet-1K benchmark.

\paragraph{Multiblock masking strategy.}
We present an extended ablation of the multiblock masking strategy where we change the targets block scale (Table~\ref{tb:ablation-multiblock-targetsize}), the context scale (Table~\ref{tb:ablation-multiblock-contexsize})  and the number of target blocks (Table~\ref{tb:ablation-multiblock-nbtargets}).
We train a ViT-B/16 for 300 epochs using \putalg with various {\tt multi-block} settings and compare performance on the 1\% ImageNet-1K benchmark using a linear probe.
In short, we find that it is important to predict several relatively large (semantic) target blocks, and to use a sufficiently informative (spatially distributed) context block.

\begin{table}[!t]
    \centering
    \begin{tabular}{l l l c}
    \multicolumn{2}{l}{\bf\small Targets} & \bf\small Context \\
    \cmidrule(lr){1-2}\cmidrule(lr){3-3}
    \small Scale & \small Freq. & \small Scale & \bf\small Top-1 \\
    \toprule
    \small (0.075, 0.2) &  4 & \small (0.85, 1.0) & 19.2 \\
    \small (0.1, 0.2) &  4 & \small (0.85, 1.0) &  39.2 \\
    \small (0.125, 0.2) &  4 & \small (0.85, 1.0) &  42.4 \\
    \small (0.15, 0.2) &  4 & \small (0.85, 1.0) & \bf 54.2 \\
    \small (0.2, 0.25) &  4 & \small (0.85, 1.0) &  38.9 \\
    \small (0.2, 0.3) &  4 & \small (0.85, 1.0) &  33.6 \\
    \end{tabular}
    \caption{\textbf{Ablation of the target block size for multi-block masking}.
    Linear evaluation on 1\% ImageNet-1K (using only 1\% of the available labels); ablating the multi-block target size during \putalg pretraining of a ViT-B/16 for 300 epochs. Predicting larger (semantic) blocks improves the low-shot accuracy as long as the context is sufficiently informative.}
    \label{tb:ablation-multiblock-targetsize}
\end{table}

\begin{table}[!t]
    \centering
    \begin{tabular}{l l l c}
    \multicolumn{2}{l}{\bf\small Targets} & \bf\small Context \\
    \cmidrule(lr){1-2}\cmidrule(lr){3-3}
    \small Scale & \small Freq. & \small Scale & \bf\small Top-1 \\
    \toprule
    \small (0.15, 0.2) &  4 & \small (0.40, 1.0) & 31.2 \\
    \small (0.15, 0.2) &  4 & \small (0.65, 1.0) & 47.1 \\
    \small (0.15, 0.2) &  4 & \small (0.75, 1.0) & 49.3 \\
    \small (0.15, 0.2) &  4 & \small (0.85, 1.0) & \bf 54.2 \\
    \end{tabular}
    \caption{\textbf{Ablation of the context size for multi-block masking}.
    Linear evaluation on 1\% ImageNet-1K (using only 1\% of the available labels); ablating the multi-block target size during \putalg pretraining of a ViT-B/16 for 300 epochs. Reducing the multi-block context size degrades the low-shot performance.}
    \label{tb:ablation-multiblock-contexsize}
\end{table}

\begin{table}[!t]
    \centering
    \begin{tabular}{l l l c}
    \multicolumn{2}{l}{\bf\small Targets} & \bf\small Context \\
    \cmidrule(lr){1-2}\cmidrule(lr){3-3}
    \small Scale & \small Freq. & \small Scale & \bf\small Top-1 \\
    \toprule
    \small (0.15, 0.2) &  1 & \small (0.85, 1.0) & 9.0 \\
    \small (0.15, 0.2) &  2 & \small (0.85, 1.0) & 22.0 \\
    \small (0.15, 0.2) &  3 & \small (0.85, 1.0) & 48.5 \\
    \small (0.15, 0.2) &  4 & \small (0.85, 1.0) & \bf 54.2 \\
    \end{tabular}
    \caption{\textbf{Ablation of the targets number for multi-block masking}.
    Linear evaluation on 1\% ImageNet-1K (using only 1\% of the available labels); ablating the multi-block number of targets during \putalg pretraining of a ViT-B/16 for 300 epochs. Increasing the number of target blocks improve the low-shot accuracy.}
    \label{tb:ablation-multiblock-nbtargets}
\end{table}

\paragraph{Masking at the output of the target-encoder.} An important important design choice in \putalg is that the target blocks are obtained by masking the \emph{output} of the target-encoder, not the input.
Table~\ref{tb:ablation-masking-output} shows the effect of this design choice on the semantic level of the learned representations when pretraining a ViT-H/16 using \putalg for 300 epochs.
 In the case where masking is applied to the input, we forward-propagate through the target-encoder once for each target region.
 Masking the output of the target-encoder during pretraining results in more semantic prediction targets and improves linear probing performance.
 \begin{table}[h]
     \centering
     \begin{tabular}{l l c c}
         \bf\small Target Masking & \bf\small Arch. & \bf\small Epochs & \bf\small Top-1 \\
         \toprule
         \cc\small Output & \cc\small ViT-H/16 & \cc300 & \cc\bf 67.3 \\
         \small Input & \small ViT-H/16 & 300 & 56.1
     \end{tabular}
     \caption{\textbf{Ablating masking output of target encoder}.
     Linear evaluation on ImageNet-1K using only 1\% of the available labels; ablating the effect of masking the target-encoder output during \putalg pretraining of a ViT-H/16 for 300 epochs.
     Masking the output of the target-encoder during pretraining significantly improves the linear probing performance of the pretrained representations.}
     \label{tb:ablation-masking-output}
\end{table}
 
\paragraph{Predictor depth.}
We examine the impact of the predictor depth on the downstream low-shot performance in Table~\ref{tb:ablation-preddepth}.
We pretrain a ViT-L/16 for 500 epochs using either a 6-layer predictor network or a 12-layer predictor network.
The model pretrained using a deeper predictor shows a significant improvement in downstream low-shot performance compared to the model pretrained with a shallower predictor.
\begin{table}[h]
     \centering
     \begin{tabular}{l l c c}
         \bf\small Predictor Depth & \bf\small Arch. & \bf\small Epochs & \bf\small Top-1 \\
         \toprule
         \small 6  & \small ViT-L/16 & 500 & 64.0 \\
         \cc\small 12 & \cc\small ViT-L/16 & \cc 500 & \cc\bf  66.9
     \end{tabular}
     \caption{\textbf{Ablating the predictor depth}.
       Linear evaluation on ImageNet-1K using only 1\% of the available labels; ablating the effect of masking the predictor depth for  a ViT-L/16 pretrained for  500 epochs.
       Increasing the predictor depth leads to significant improvement of the linear probe performance of the pretrained representations.}
     \label{tb:ablation-preddepth}
\end{table}

\paragraph{Weight decay.} In Table~\ref{tb:ablation-weightdecay}, we evaluate the impact of weight-decay during pretraining. We explore two weight decay strategies: linearly increase the weight-decay from $0.04$ to $0.4$ or use a fix weight-decay of $0.05$. Using a smaller weight decay during pretraining improves the downstream performance on ImageNet-1\% when fine-tuning. However, this also leads to a degradation of performance in linear evaluation. In the main paper, we use the first weight decay strategy as it improves the performances in linear evaluation downstream tasks.

\begin{table}[h]
     \centering
     \begin{tabular}{l l c c c}
         \bf\small Weight Decay & \bf\small Arch. & \bf\small Epochs & \bf\small ImageNet-1\% & \bf ImageNet Linear-Eval \\
         \toprule
         \cc \small $0.04 \rightarrow 0.4$  & \cc\small ViT-L/16 &\cc 600 &  \cc 69.4 & \cc\bf 77.8  \\
         \small $0.05$ & \small ViT-L/16 &  600 & \bf  70.7 & 76.4\\
     \end{tabular}
     \caption{\textbf{Ablating the pretraining weight-decay}.
       We compare our default pretraining  weight decay strategy where we linearly increase the weight-decay from $0.04$ to $0.4$ to using a fix weight decay of $0.05$. Using a smaller weight-decay during pretraining can improve the fine-tuning performance on ImageNet-1\%, However, it also leads to a drop of performance in linear evaluation.}
     \label{tb:ablation-weightdecay}
\end{table}

\paragraph{Predictor width.} We explore the impact of the predictor width in Table~\ref{tb:ablation-predwidth}. We compare \putalg using a ViT-L encoder and a predictor with  $386$ channels to a similar  model using a predictor with $1024$ channels. Note that the ViT-L encoder has  $1024$ channels. Using a bottleneck in the predictor width improves the downstream performance on ImageNet 1\%.

\begin{table}[h]
     \centering
     \begin{tabular}{l l c c}
         \bf\small Predictor Width & \bf\small Arch. & \bf\small Epochs & \bf\small Top-1 \\
         \toprule
         \cc\small 384  & \cc\small ViT-L/16 & \cc 600 & \cc\bf 70.7 \\
         \small 1024 & \small ViT-L/16 &  600 &   68.4
     \end{tabular}
     \caption{\textbf{Ablating the predictor width.} We reports results on  ImageNet-1K 1\% using fine-tuning. We compare two predictors having a width of either $384$ or $1024$. Note the \putalg encoder is a ViT-L with  $1024$ channels. Having a width bottleneck in the predictor improves the  downstream performances.}
     \label{tb:ablation-predwidth}
\end{table}

\section{Finetuning on the full ImageNet}
In this section, we report performance on \putalg when fine-tuning on the full ImageNet dataset. We focus on the ViT-H/16$_{448}$ as this architecture achieves state-of-art performance with MAE~\cite{he2021masked}.

We use a fine-tuning protocol similar to MAE. Specifically, we fine-tune our model for $50$ epochs using AdamW and a cosine learning rate schedule. The base learning rate is set to $10^{-4}$ and the batch size to $528$. We train using  mixup~\cite{zhang2018mixup} set to $0.8$, cutmix~\cite{yun2019cutmix} set to $1.0$, a drop path probability of $0.25$ and a weight decay set to $0.04$. We also use a layer decay of $0.75$. Finally, we use the same rand-augment data-augmentations as MAE,

Table~\ref{tb:ablation-finetune} reports the fine-tuning results. \putalg achieves $87.1$ top-1 accuracy. Its performance is less than $1\%$ away from the best MAE model despite \putalg being trained for $5.3$ times less epochs than MAE.
This result demonstrates that \putalg is competitive when fine-tuning on the full ImageNet dataset.
\begin{table}[h]
     \centering
     \begin{tabular}{l l c c}
         \bf\small Method & \bf\small Arch. & \bf\small Epochs & \bf\small Top-1 \\
         \toprule
         \small MAE~\cite{he2021masked}  & \small ViT-H/14$_{448}$ & 1600 & \bf 87.8 \\
         \cc\small \putalg & \cc\small ViT-H/16$_{448}$ &  \cc300 &   \cc 87.1
     \end{tabular}
     \caption{\textbf{Finetuning on the full ImageNet dataset}.
       \putalg achieves competitive performance. \putalg is close to MAE approach despite \putalg being trained for $5.3$ times less epochs than MAE.}
     \label{tb:ablation-finetune}
\end{table}

\section{RCDM Visualizations}

To visualize the representations of a pretrained neural network in pixel space, we use the RCDM framework~\cite{bordes2022high}.
The RCDM framework trains a decoder network $h_\omega$, comprising a generative diffusion model, to reconstruct an image $\vx$ from the representation vector of that image $\vs_x$ and a noisy version of that image $\hat{\vx} \coloneqq \vx + \epsilon$, where $\epsilon$ is an additive noise vector.
Concretely, the decoder objective is to minimize the loss function $\lVert h_\omega(\hat{\vx}, \vs_x) - \epsilon \rVert$.
We train each RCDM network for 300,000 iterations 
using the default hyperparameters~\cite{bordes2022high}.
After training the decoder, one can subsequently feed the representation vector of an unseen test image $\vs_y$ into the decoder along with various random noise vectors to generate several pixel-level visualizations of the representation, thus providing insight into the features captured in the representations of the pretrained network.
Qualities that are common across samples represent information that is contained in the representation.
On the other hand, qualities that vary across samples represent information that is not contained in the representations

In Figure~\ref{fig:visualization-predictor}, the visualizations are obtained by feeding the average-pooled output of the predictor, conditioned on a specific target region, into the decoder network, along with various random noise vectors.
In Figures~\ref{fig:visualization-encoder} and~\ref{fig:visualization-msn}, the visualizations are obtained by feeding the average-pooled output of the target-encoder into the decoder network, along with various random noise vectors.

\subsection{Encoder Visualization}

\begin{figure*}[h]
    \centering
    \includegraphics[height=0.35\textheight]{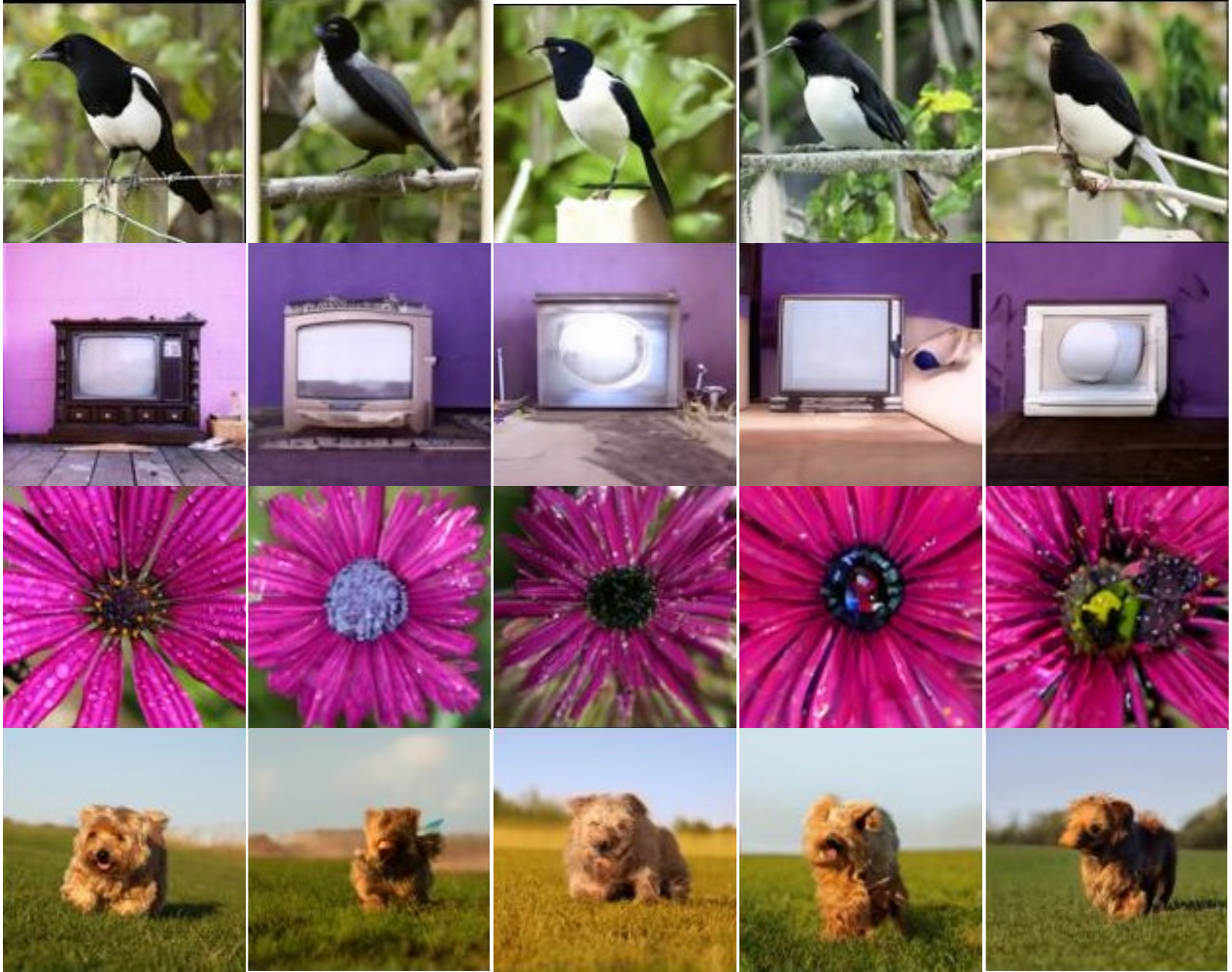}
    \caption{\textbf{Visualization of \putalg target-encoder representations}.
        For each image: first column contains the original image; subsequent columns contain samples from a generative model decoding the average-pooled output of a pretrained \putalg target-encoder.
        Qualities that are common across samples represent information that contained is in the \putalg representation. \putalg is able to correctly capture the high-level information regarding objects and their poses.
        Qualities that vary across samples represent information that is not contained in the representation. \putalg encoder discards the precise low-level details as well as background information.}
    \label{fig:visualization-encoder}
\end{figure*}

In Figure~\ref{fig:visualization-encoder}, we visualize the average-pooled \putalg representations at the output of our ViT-H/14 target-encoder.
The first column contains the original image, while subsequent columns contain synthetic samples obtained by feeding the average-pooled representation of the image into the decoder along with various random noise vectors.
Figure~\ref{fig:visualization-encoder} suggests that   the \putalg target-encoder is able to correctly capture the high-level information regarding objects and their poses, while discarding low-level image details and background information.

\begin{figure*}[h]
    \centering
    \includegraphics[height=0.35\textheight]{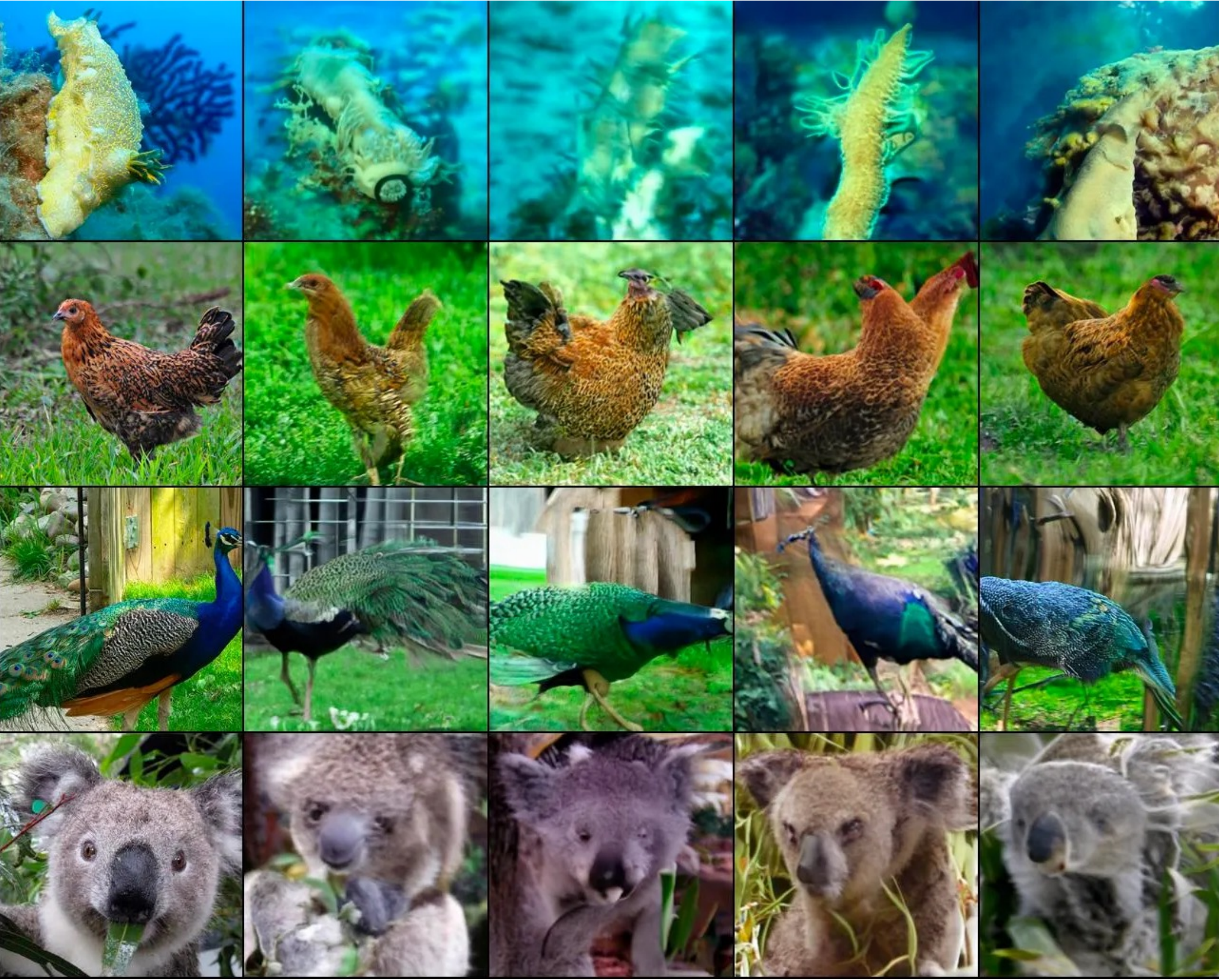}
    \caption{\textbf{Visualization of MSN target-encoder representations}.
    For each image: first column contains the original image; subsequent columns contain samples from a generative model decoding the  output of a frozen MSN encoder~\cite{assran2022masked}.
    Qualities that are common across samples represent information that is contained in the representation. Qualities that vary across samples represent information that is not captured by MSN. Compared to \putalg, MSN samples show higher variability. MSN retains less information from the input. In particular, it discards global structure information such as the  object pose or even number of instances.} 
\label{fig:visualization-msn}
\end{figure*}

Figure~\ref{fig:visualization-msn} shows similar visualizations, but when using an MSN~\cite{assran2022masked} pretrained ViT-L/7 target-encoder to compute the image representations.
The MSN method trains a context- and target-encoder using a Joint-Embedding Architecture to enforce invariance of global image representations to various hand crafted data augmentations and missing patches.
While the MSN pretrained network is able to capture high level semantic information about the image in the first column, it also exhibits higher variability in the generated samples, e.g., variability in the object pose, object scale, and number of instances.
In short, the MSN pretrained discards much of the local structure in the image, which is in stark contrast to \putalg, which retains information about much of the local structure in the input image.
\pagebreak


\end{document}